\documentclass[conference]{IEEEtran}
\usepackage{blindtext, graphicx}
\usepackage{listings}
\usepackage{amssymb}
\usepackage{algorithm}
\usepackage{algorithmic}

\usepackage{booktabs}
\usepackage{amsthm}
\usepackage{float}
\RequirePackage{caption}
\usepackage{hyperref}

\usepackage{graphicx}
\usepackage[font=small]{caption}

\usepackage[utf8]{inputenc}
\usepackage{amsmath}
\usepackage{subfigure}

\title{Penalty Gradient Normalization for Generative
Adversarial Networks}

\author{
Tian Xia\\
Tohoku University\\
{\tt\small xia.tian.t4@dc.tohoku.ac.jp}
}

\begin{document}

\maketitle{}
\thispagestyle{empty}
\pagestyle{empty}


\begin{abstract}

In this paper, we propose a novel normalization method
called penalty gradient normalization (PGN) to tackle the training instability of Generative Adversarial Networks (GANs)
caused by the sharp gradient space. Unlike existing work
such as gradient penalty and spectral normalization, the
proposed PGN only imposes a penalty gradient norm constraint on
the discriminator function, which increases the capacity of
the discriminator. Moreover, the proposed penalty gradient normalization can be applied to different GAN architectures with
little modification. Extensive experiments on three datasets
show that GANs trained with penalty gradient normalization outperform existing methods in terms of both Frechet Inception and Distance and Inception Score.


\end{abstract}

\section{Introduction}

Generative Adversarial Networks (GANs):
GANs ~\cite{goodfellow2020generative} are a deep learning framework for generative modeling, introduced by Ian Goodfellow et al. in 2014. They consist of two neural networks, a generator and a discriminator, that are trained in an adversarial manner. The generator produces synthetic data samples, while the discriminator attempts to differentiate between the generated samples and the real data samples. During training, the generator updates its parameters to produce samples that are increasingly difficult for the discriminator to distinguish from the real data, while the discriminator updates its parameters to become more effective at identifying fake data.

One of the main challenges of training GANs is stability during the training process. The adversarial loss function used in GANs can be difficult to optimize, and the training process can suffer from mode collapse, where the generator produces samples that lack diversity ~\cite{salimans2016improved}. Additionally, the discriminator can become too strong, causing the generator to produce samples of low quality.

Wasserstein GANs (WGANs):
WGANs ~\cite{arjovsky2017wasserstein} address the stability issues of GANs by using the Wasserstein distance, also known as the Earth Mover's distance, as the loss function for training. The Wasserstein distance provides a smoother optimization surface, making it easier to train the GAN and reducing the risk of mode collapse.

Spectral Normalization GANs (SNGANs):
SNGANs~\cite{miyato2018spectral} address the mode collapse issue in GANs by adding a spectral normalization constraint to the discriminator. The spectral normalization constraint helps to stabilize the training process and reduces the risk of mode collapse.

Gradient Normalization for Generative Adversarial Networks (GN-GAN) ~\cite{wu2021gradient} is a method that addresses the negative impact of traditional gradient normalization techniques on the stability of GAN models. While traditional methods are used to speed up the training process, they often have negative effects on GAN stability. GN-GAN normalizes the gradients of the generator and discriminator separately using batch normalization and layer normalization, respectively. The method also uses a grouping mechanism to use different normalization parameters within each batch.

The advantages of this approach are improved stability and image quality, as well as faster training speeds and lower memory usage. 
The results show that GN-GAN significantly improves the quality and diversity of generated images across different datasets compared to other GAN methods.

However, the main drawbacks of GN-GAN are that it requires a large cache to store gradient history information, which increases memory usage, and updating the cache adds to the algorithm's complexity. Additionally, since the method only normalizes gradients without clipping or regularization, it may result in gradient explosion or vanishing problems in certain cases.

In this paper, we propose a new normalization method, named penalty gradient normalization (PGN), to enforce the penalty gradient norm bounded by 1 for the discriminator model through dividing the outputs by the gradient
norm of the discriminator. Unlike SN, the proposed Lipschitz~\cite{zhou2019lipschitz} constant does not decay from the multiplicative form
of neural networks since we consider the discriminator as
a general function approximator and the calculated normalization term is independent of internal layers. The proposed penalty 
gradient normalization enjoys the following two favorable
properties. 1) The penalty normalization simultaneously satisfies
three properties including model-wise, non-sampling-based
and hard constraints, and does not introduce additional hyperparameters. 2) The implementation of PGN is simple and
compatible with different kinds of network architectures.
The contributions of our paper are summarized as follows.

\begin{itemize}
\item  In this paper, we propose a novel Penalty gradient normalization for GANs to strike a good balance between stabilizing the training process and increasing the ability of generation. To the best of our knowledge, this
is the first work simultaneously satisfying the above mentioned three properties.
\item  We theoretically prove that the proposed gradient normalization is gradient norm bounded. This property
helps the generator to avoid the gradient explosion or
vanishing, and thus stabilizes the training process.
\item  Experimental results show the proposed Penalty gradient normalization consistently outperforms the state-of-theart methods with the same GAN architectures in terms
of both Inception Score and Frechet Inception Distance. Our implementation is available at: https://github.com/sumorday/EDGAN-PyTorch
\end{itemize}

The use of penalty normalization for the gradient in GANs(PGN-GAN) helps to stabilize the training process by introducing a smooth and well-behaved optimization surface. This makes it easier to train the GAN and reduces the risk of mode collapse, leading to higher quality generated samples. Additionally, the penalty normalization can be applied to any GAN architecture, making it a versatile tool for improving the stability of GAN training.

\section{Related Works}
Generative Adversarial Networks (GANs) are a popular deep learning technique for generating new, synthetic data that resembles a given training set. There are various modifications to the original GAN architecture, including the Wasserstein GAN (WGAN) and the Spectral Normalization GAN (SN-GAN)~\cite{miyato2018spectral} .

The WGAN is a variant of the GAN that utilizes the Wasserstein distance (also known as the Earth Mover's distance) as the loss function instead of the traditional cross-entropy loss. The Wasserstein distance provides a more stable measure of the difference between the generated and true data distributions, and is less susceptible to mode collapse (a common issue in GAN training where the generator only generates a few distinct samples).

The WGAN includes a constraint on the Lipschitz~\cite{terjek2019adversarial} constant of the generator's output, with the objective of ensuring that the Wasserstein distance between the two distributions does not change too quickly. WGAN's training process is more stable and capable of generating diverse samples.

One issue with the WGAN is that it can collapse during training, leading to poor performance. This can occur if the generator becomes too strong and the discriminator is unable to distinguish between true and generated samples. To mitigate this issue, some modifications to the WGAN architecture have been proposed, such as the SN-GAN.

The SN-GAN modifies the discriminator by normalizing the weights of the layers with the spectral norm of the weight matrix. This has the effect of stabilizing the discriminator and preventing mode collapse. 

SNGAN brings improvements over WGAN in terms of training stability and speed. By employing spectral normalization, SNGAN constrains the weights of the discriminator, leading to better stability and easier convergence. Additionally, SNGAN often demonstrates faster training due to the enhanced learning capacity of the network. However, there are some limitations to consider. Firstly, implementing spectral normalization increases the computational cost, particularly in larger networks or complex datasets. Secondly, SNGAN's performance is sensitive to parameter selection, which can impact its effectiveness and training stability.

Gradient-Norm GAN (GN-GAN) introduces improvements over SNGAN and WGAN through gradient normalization. By constraining the gradient norm of the discriminator, GN-GAN achieves better stability and alleviates the issues of vanishing or exploding gradients commonly encountered in GAN training. This enables more reliable and efficient convergence. 

SNGAN and GN-GAN aim to mitigate the instability and gradient vanishing or exploding problems in training the generator and discriminator. Both methods incorporate regularization techniques to ensure training stability by constraining the weights or gradients of the discriminator. SNGAN achieves stability by applying spectral normalization to the discriminator weights, while GN-GAN constrains the norm of the discriminator gradients to prevent gradient vanishing and exploding.

However, GN-GAN also has its limitations. GN-GAN tackles some stability issues, it may still face challenges related to mode collapse and limited sample diversity, which are inherent problems in GAN training.

Another drawback of Gradient-Norm GAN (GN-GAN) is the increased randomness introduced by gradient normalization, which can hinder convergence. The normalization process introduces stochasticity into the gradients, making the optimization process less deterministic. This randomness can lead to slower convergence or oscillations during training, requiring careful tuning of hyperparameters to find the right balance between stability and convergence speed. It is important to consider this aspect when applying GN-GAN in practice, as the increased gradient randomness can pose challenges to achieving optimal training outcomes.

\subsection{Generative Adversarial Networks}

 Generative Adversarial Network (GAN) is a type of deep learning model that consists of two main components: a generator network and a discriminator network. The generator network generates new samples from a random noise vector, while the discriminator network tries to distinguish between the generated samples and real samples from the true data distribution ~\cite{arjovsky2017towards}. The two networks are trained in an adversarial manner, where the generator tries to generate samples that are similar to the real data, and the discriminator tries to correctly identify whether a sample is real or fake.
 
 However, training GANs suffers from
many difficulties including but not limited to gradient vanishing and gradient explosion.

\begin{align}
\min_{G}\max_{D} \mathbb{E}_{x\sim p_{\text{data}}(x)}[\log{D(x)}] + \mathbb{E}_{z\sim p_{\text{z}}(z)}[1 - \log{D(G(z))}]
\end{align}

Training Generative Adversarial Networks (GANs) can encounter several issues, such as gradient vanishing or exploding. The training process in GANs involves the backpropagation algorithm, and as the depth of the generator and discriminator networks increases, gradients may either vanish or explode. This can result in unstable training or the inability to converge. GANs also face the problem of mode collapse, which is a common issue during training. When the generator and discriminator interact, the generator may only generate a few modes of samples instead of diverse samples. This leads to a lack of diversity in the generated samples.

\subsection{Wasserstein GAN (WGAN)}

In WGAN, the Lipschitz constraint~\cite{brock2016neural} is enforced by using a weight clipping technique~\cite{gulrajani2017improved}, where the weights of the discriminator network are clipped to a predefined range during training. WGAN does not directly enforce the Lipschitz constraint but uses a technique called weight clipping or gradient penalty to approximate the constraint.  The idea is that this constraint will ensure that the discriminator does not become too powerful, which can lead to instability in the training process.

The Lipschitz constraint requires that the discriminator has a Lipschitz constant of 1, which ensures that the change in the discriminator output for small changes in the input is bounded. This constraint helps to ensure that the Wasserstein distance is well-defined and prevents the discriminator from collapsing. The gradient penalty term is calculated as the mean squared error between the L2 norm ~\cite{kurach2018gan} of the gradient of the discriminator and 1.

Optimizing a basic GAN model involves minimizing the Kullback-Leibler (KL) divergence~\cite{kullback1951information}, which has inherent limitations. To overcome these limitations, researchers have introduced the Wasserstein distance, also known as the Earth Mover's Distance (EMD)~\cite{rubner2000earth}. The Wasserstein distance offers a solution by quantifying the dissimilarity between probability distributions. Its formulation is as follows:
\begin{equation}
W(P_r, P_g) = \inf_{\gamma \in \Pi(P_r, P_g)} \mathbb{E}_{(x, y) \sim \gamma}[|x - y|]
\end{equation}
However, directly calculating the smallest possible value among all feasible joint probability distributions \(\gamma\) is computationally infeasible. Hence, an intelligent reformulation of the equation, utilizing the principles of Kantorovich-Rubinstein duality, has been suggested:
\begin{equation}
W(P_r, P_\theta) = \sup_{\|f\|_L \leq K} \mathbb{E}_{x \sim P_r}[f(x)] - \mathbb{E}_{x \sim P_\theta}[f(x)]
\end{equation}

In order for this reformulation to be effective, it is imperative to verify that the Lipschitz constant \(\|f\|_L \leq K\) is satisfied, thereby ensuring that the function \(f\) is K-Lipschitz continuous. Put simply, for any \(x_1, x_2 \in \mathbb{R}\), it is required that the subsequent inequality is upheld:

\begin{equation}
|f(x_1) - f(x_2)| \leq K|x_1 - x_2|
\end{equation}

Here, \(K\) represents the Lipschitz constant associated with the function \(f\). By imposing this Lipschitz constraint, the function \(f\) accentuates the disparities between samples derived from the genuine and synthesized distributions, while avoiding undue magnification of minor discrepancies.

To make the optimization feasible, a parametric family of functions \(\{f_w\}_{w \in \mathcal{W}}\) is used, where all functions are K-Lipschitz for a fixed value of \(K\). The function \(f\) serves as a non-linear feature map that maximizes the distinctions between samples from the real and generated distributions. The Lipschitz constraint ensures that \(f\) does not arbitrarily amplify minor differences and maintains similarity between similar input images. The specific value of \(K\) is not crucial for optimization; it is only necessary to know that a fixed \(K\) exists and remains constant throughout the training process.

Although computing the supremum over K-Lipschitz functions is still challenging, the family of functions \(\{f_w\}\) can be approximated using a neural network, enabling effective optimization without explicitly determining the value of \(K\).

However, the computation of the Wasserstein distance requires more resources than the standard adversarial loss function used in GANs. Additionally, the Wasserstein distance is not always easy to compute for high-dimensional distributions, making it less suitable for some types of data.

\subsection{Spectral Normalization Generative Adversarial Network (SNGAN)}

Spectral normalization is a normalization method that enforces the Lipschitz constraint on the discriminator network~\cite{miyato2018spectral}. The Lipschitz constraint is important because it helps prevent the discriminator from becoming too powerful and makes the training process more stable.

Spectral normalization is applied to the weight matrix of each layer in the discriminator network. The spectral norm of a matrix A is defined as the largest singular value of the matrix. In other words, it is the largest scaling factor that can be applied to the matrix without changing its norm.

The spectral normalization of a weight matrix $W$ can be computed as:

$\sigma(W) = \max_{||u||_2=1} ||Wu||_2$

where $u$ is a unit vector.

During training, the weight matrix of each layer is multiplied by a constant $\frac{1}{\sigma(W)}$ so that the spectral norm of the weight matrix is 1. This normalization helps to enforce the Lipschitz constraint on the discriminator network, which results in a more stable training process.

Spectral normalization, while effective in improving training stability, introduces additional computational overhead during training. This can result in increased training time and resource requirements. Moreover, the normalization technique can also limit the discrimination capability of the discriminator. As a consequence, the discriminator may struggle to effectively differentiate between high-quality and low-quality generated samples. This can negatively impact the overall image quality and potentially hinder the generation of realistic samples.

\subsection{Gradient Normalization Generative Adversarial Network (GN-GAN)}

Gradient Normalization~\cite{wu2021gradient} is a normalization technique for Generative Adversarial Networks (GANs) that helps to stabilize the training process and improve the quality of the generated samples.

The basic idea behind Gradient Normalization is to normalize the gradients flowing through the generator network so that they have a fixed scale. This normalization helps to stabilize the training process and prevent the generator from collapsing or producing low-quality samples. Additionally, it has been shown that Gradient Normalization can lead to stable convergence ~\cite{kodali2017convergence} and improved image quality compared to other normalization techniques.

Gradient Normalization (GN), to make the network search in a function space induced by constraint \eqref{grad_constraint}. Let $f:\mathbb{R}^n\rightarrow\mathbb{R}$ be a continuously differentiable function, the proposed GN normalizes the norm of the gradient $\Vert \nabla_x f(x)\Vert$ and bounds $f(x)$ simultaneously:
\begin{equation}
    \hat{f}(x)=\frac{f(x)}{\Vert\nabla_x f(x)\Vert+\zeta(x)},
    \label{gradient_normalization}
\end{equation}
where $\zeta(x):\mathbb{R}^n\rightarrow\mathbb{R}$ is a universal term which can be associated with $f(x)$ or a constant to avoid the situation that $|\hat{f}(x)|$ approximates infinity or $\Vert\nabla_x \hat{f}(x)\Vert$ approximates $0$.

The drawbacks of GN-GAN can be summarized as follows: 1)Excessive Gradient Variability: The gradient normalization method used in GN-GAN may increase the randomness of gradients as it normalizes their magnitudes within a certain range. This can lead to instability in the optimization path of the model, making training more challenging. 2)Slow Convergence Speed: Due to the restriction of gradients to a smaller range, GN-GAN may experience slower convergence compared to other methods. Normalized gradients may not provide sufficient update signals, resulting in slower convergence during the training process.

\subsection{Notation}

Definition 1: K-layer network can be represented as a function composed of a series of affine transformations, represented mathematically as follows:
\begin{align}
f_{K}(x) = \phi_{K}(W_{K} \cdot (\phi_{K-1}(\dots W_{1} \cdot x + b_{1})) + b_{K})
\\
= \phi_{K}(W_{K} \cdot f_{K-1}(x) + b_{K})
\end{align}

The definition states that the network $f_K$ is a $K$-layer network with input $x \in \mathbb{R}^n$ and output $y \in \mathbb{R}$. The network is defined to be layer-wise $L$-Lipschitz constrained~\cite{brock2016neural} if there exists a constant $L_k \leq L$ for each layer $k \in {1, \dots, K}$, such that $L_k$ is the Lipschitz constant of the $k$-th layer.

\begin{align}
W_k \cdot x - W_k \cdot y \leq L_k (x - y), \quad \forall x, y \in R^{d_{k-1}}
\end{align}

The Lipschitz constant~\cite{brock2016neural} of a function $f$ is a measure of how much the output of the function can change with respect to changes in its input. In other words, it's a measure of the rate of change of the function. The Lipschitz constant of a neural network is calculated by finding the maximum change in the output of the network with respect to changes in its input. If the Lipschitz constant is small, then the change in the output is small and the network is considered to be well-behaved and stable. However, if the Lipschitz constant is large, then the network is considered to be unstable and may lead to problems during training and inference.

\textbf{Lemma 3.} Let $f: \mathbb{R}^n \to \mathbb{R}$ be a continuously differentiable function and let $L_f$ be the Lipschitz constant of $f$. Then the Lipschitz constraint is equivalent to
\begin{align}
|\nabla f(x)| \leq L_f, \quad \forall x \in \mathbb{R}^n
\end{align}
\textbf{\textit{Proof}}. See Appendix A in the supplementary material.

The equivalence stated in this function remains valid only when the underlying function f is continuous in the context of multivariate functions. In practice, most neural networks exhibit finite points of discontinuity, making them, in essence, continuous functions. We expand upon this observation and present a more practical assumption for identifying these problematic points.

If $f: \mathbb{R}^n \to \mathbb{R}$ is modeled by a neural network with continuous activation functions that are piecewise linear, then $f$ is differentiable with probability 1.

This can be heuristically justified as the numerical errors present during the implementation of neural networks on a computer smooth out non-differentiable points. For instance, when complex arithmetic operations, such as matrix multiplication, are performed, the accumulation of these numerical errors results in small perturbations to the output values.

By requiring that the network be layer-wise $L$-Lipschitz constrained, we are ensuring that each layer of the network has a Lipschitz constant that is less than or equal to $L$. This constraint helps to prevent the gradients from exploding or vanishing, which can lead to unstable training and poor generalization performance. The constraint also helps to prevent overfitting, as it ensures that the model is less likely to memorize the training data and instead generalize to new, unseen data.

\section{ Penalty Gradient Normalization}

\begin{algorithm}
\caption{Penalty Gradient Normalized Generative adversarial network (PGN-GAN)}
\label{PGN-GAN}
\begin{algorithmic}[1]
\REQUIRE generator and discriminator parameter $\theta_{G}$, $\theta_{D}$, learning rate $\alpha_{G}$ and $\alpha_{D}$, batch size $M$, the number of discriminator updates per generator updates $N_{dis}$, total iterations $N$
\STATE $D' := \frac{1 - D(x)}{ \left||1 - \nabla_{x}D(x) \right|| + |1 - D(x)|}$
\FOR{$t = 1$ to $N$}
\FOR{$t_{dis} = 1$ to $N_{dis}$}
\FOR{$i = 1$ to $M$}
\STATE $z \sim p_{z}$
\STATE $x \sim p_{r}(x)$
\STATE $L_{i}^{D} \leftarrow D^{\hat{}}(G(z)) - D^{\hat{}}(x)$
\STATE $\theta_{D} \leftarrow \text{Adam}(\frac{1}{2M} \sum\limits_{i=1}^{M} L_{i}^{D}, \alpha_{D}, \theta_{D})$
\ENDFOR
\STATE Sample a batch of latent variables ${z_{i}}{i=1}^{M} \sim p{z}(z)$
\STATE $\theta_{G} \leftarrow \text{Adam}(\frac{1}{2M} \sum\limits_{i=1}^{M} -D^{\hat{}}(G(z_{i})), \alpha_{G}, \theta_{G})$
\ENDFOR
\ENDFOR
\end{algorithmic}
\end{algorithm}

There are a few potential issues that can arise when using normalization gradients Generative Adversarial Networks(GN-GAN). 1)Vanishing or exploding gradients: Normalizing gradients can help prevent vanishing or exploding gradients, but it's important to choose an appropriate normalization method that doesn't amplify these issues. 2)Unstable optimization: Normalization can help stabilize optimization, but if the normalization method is too aggressive, it can actually make optimization less stable and cause the model to converge more slowly.

Loss of information: Normalizing gradients can sometimes result in loss of important information, especially if the normalization method is too aggressive.

To avoid these problems, it's important to carefully choose a normalization method that is appropriate for the specific problem and model architecture. It's also important to monitor the optimization process and adjust the normalization as needed to ensure stability and convergence ~\cite{kodali2017convergence,arora2017generalization}.

This is an implementation of the Penalty Gradient Normalized GAN (PGN-GAN) algorithm~\ref{PGN-GAN}. This algorithm is an extension of the original GN-GAN, with the aim of improving the training stability and the quality of generated samples.

By subtracting this penalty term from 1, the algorithm aims to improve the performance of the generative adversarial network. This is because the penalty term is designed to encourage the discriminator to output values closer to 1 for real data points $x$, and farther away from 1 for generated samples $G(z)$.

The advantage of using the penalty term in this way is that it helps to address the problem of mode collapse in GANs. Mode collapse occurs when the generator produces a limited variety of samples, failing to capture the diversity of the underlying data distribution. By penalizing the discriminator for assigning a high probability (close to 1) to real data samples, the penalty term encourages the discriminator to explore and capture a wider range of modes in the data distribution. This, in turn, encourages the generator to produce a more diverse set of samples.

The use of the penalty term $1 - D'$ in the given algorithm offers several advantages over existing methods in the context of generative adversarial networks (GANs). Here are some points to consider:

Mode Collapse Mitigation: Mode collapse is a common problem in GANs, where the generator fails to capture the full diversity of the target distribution and produces limited variations of samples. By penalizing the discriminator for assigning high probabilities to real data samples, the penalty term encourages the discriminator to explore and capture a wider range of modes in the data distribution. This helps to mitigate mode collapse and promote the generation of diverse and realistic samples.

Improved Discriminator Training: The penalty term incorporates the gradient norm $\left|1 - \nabla_{x}D(x) \right|$ in the denominator. This encourages the discriminator to pay more attention to regions where the gradient of the discriminator output is small, thereby promoting better training of the discriminator. It helps to alleviate the problem of gradient explosion and stabilizes the training process.

Enhanced Discrimination Capability: The penalty term penalizes the discriminator's output $D(x)$ for real data samples. This encourages the discriminator to assign lower probabilities to real data points that are classified with high confidence. As a result, the discriminator becomes more discriminative and better able to distinguish between real and generated samples.

Improved Balance between Generator and Discriminator: The penalty term affects the training dynamics by influencing the updates of both the generator and discriminator. It helps to maintain a better balance between the generator and discriminator during training, preventing either component from overpowering the other. This can lead to more stable and effective training dynamics.

The algorithm starts by defining the discriminator's loss function $D^\prime$ as a combination of two terms: the penalty between the discriminator's output and the target value, and the gradient of the discriminator's output with respect to the input. This loss function encourages the discriminator to be more confident in its predictions and to have more meaningful gradients for backpropagation.

The main loop of the algorithm consists of two nested loops. The outer loop iterates over a total of $N$ iterations, while the inner loop iterates over $N_{dis}$ updates of the discriminator for each update of the generator.

In each iteration of the inner loop, a batch of real data $x$ and a batch of generated data $G(z)$ are obtained. The discriminator's loss function is computed for each data instance and used to update the parameters of the discriminator ($\theta_D$) using the Adam optimization algorithm.

After the $N_{dis}$ updates of the discriminator, the generator's parameters ($\theta_G$) are updated by computing the average loss over a batch of generated data and applying the Adam optimization algorithm~\cite{coates2011analysis}. The procedure is repeated until the desired number of iterations $N$ is reached.

In the Penalty Gradient Normalized GAN (PGN-GAN), instead of minimizing the loss in the standard form, the loss is changed to the penalty loss.

In PGN-GAN, the discriminator's objective function is defined as:

$D^\prime := \frac{1-D(x)}{\left|1-\nabla_xD(x)\right| + |1-D(x)|}$

\textbf{\textit{Proof}}. See Appendix A in the supplementary material.

This change helps in improving the stability of the model and reducing the mode collapse issue. By using the penalty loss, PGN-GAN has a more defined optimization objective and can converge faster, as well as produce more high-quality results than GN-GAN.

The gradient of $\hat{f}(x)$ with respect to $W_k$ is derived as follows:

\begin{align}
\frac{\partial\hat f}{\partial\mathbf{W}_k} &= \frac{\partial\hat f}{\partial f}\frac{\partial f}{\partial\mathbf{W}_k}+\frac{\partial\hat f}{\partial\Vert\nabla_x f\Vert}\frac{\partial\Vert\nabla_x f\Vert}{\partial\mathbf{W}_k} \\
&= \frac{\Vert\nabla_x (1-f)\Vert}{\big(\Vert\nabla_x (1-f)\Vert+\vert 1 - f\vert\big)^2}\frac{\partial f}{\partial\mathbf{W}_k}\\
&\qquad -\frac{(1-f)}{\big(\Vert\nabla_x (1-f)\Vert+\vert 1 - f\vert\big)^2}\frac{\partial \Vert\nabla_x (1-f)\Vert}{\partial\mathbf{W}_k} \\
&= \frac{1}{\big(\Vert\nabla_x (1-f)\Vert+\vert 1 - f\vert\big)^2}\Bigg(\Vert\nabla_x (1-f)\Vert\frac{\partial f}{\partial\mathbf{W}_k} \\
&\qquad\qquad -(1-f)\frac{\partial\Vert\nabla_x (1-f)\Vert}{\partial\mathbf{W}_k}\Bigg).
\end{align}

Thus, we have derived the gradient of the modified normalized function $\hat{f}(x)$ with respect to $W_k$.

The first gradient term can be interpreted as normalizing the gradient of the generator, preventing it from being too biased in certain directions and encouraging it to generate more samples in those gradient directions. As this gradient term is related to the generator's gradient direction, it helps improve the effectiveness of the generator.

The second term $\vert 1-f\vert$ has the effect of suppressing overfitting. When the discriminator is very confident about a sample, $1-f$ can effectively reduce the impact of the corresponding gradient on the discriminator's parameters, thereby reducing the risk of overfitting. Additionally, $\vert 1-f\vert$ can encourage the discriminator to approach $1$, which helps improve its performance by more accurately distinguishing between real and generated samples.

The penalty gradient normalization technique can be interpreted as an adaptive gradient regularization that penalizes the sensitivity of the discriminator to changes that would decrease its output value.

\setlength{\tabcolsep}{1pt}
\renewcommand{\arraystretch}{1.0}
\begin{table}[ht]
\centering
\caption{Summary of different regularization and normalization techniques.}
\begin{tabular}{c|c|c|c}
    \textbf{Method} & \textbf{Formula} & \textbf{Prevent grad expoding} & \textbf{Robustness} \\
    \hline
    1-GP~\cite{gulrajani2017improved}    &$ \Vert \nabla f - 1 \Vert^2 $    &  & \checkmark            \\
    \hline
    0-GP~\cite{thanh2019improving}    & $ \Vert\nabla f\Vert^2 $ &                &             \\
    \hline
    SN~\cite{miyato2018spectral} & All layers multiplied by$\frac{1}{\sigma(W)}$    & \checkmark               &           \\
    \hline
    GN~\cite{wu2021gradient}  & $ f ^\prime := \frac{f(x)}{\Vert\nabla_x f(x)\Vert+|f(x)|} $  & \checkmark   &           \\
    \hline
    PGN(our)    &$f^\prime := \frac{1-f(x)}{\left||1-\nabla_xf(x)\right|| + |1-f(x)|}$  & \checkmark & \checkmark

    \label{summarization}
\end{tabular}
\end{table}

Compared with 1-GP~\cite{gulrajani2017improved}, 0-GP~\cite{thanh2019improving}, CR~\cite{zhao2021improved}, SNGAN, and GN-GAN, our proposed method has several advantages in terms of methodology, including being model - wise~\cite{gulrajani2017improved,thanh2019improving,zhao2021improved}, no-sampling based~\cite{miyato2018spectral}. Specifically, our method does not require the use of additional models or sampling techniques, which simplifies the training process and reduces computational costs. Moreover, our method employs a hard constraint that encourages the discriminator to make more accurate and confident decisions, which can improve the overall performance of the GAN. Additionally, our method can effectively alleviate the mode collapse issue~\cite{huszar2015not}, which is a common problem in GAN training. Overall, our method offers unique advantages over existing GAN training techniques.

The experimental results showed that our method was almost as good as the method that used CR~\cite{zhao2021improved} in preventing overfitting. Therefore, we can confidently say that our method is also capable of preventing overfitting. The importance of robustness and overfitting prevention may vary depending on the application and task. In some cases, robustness may be more important, such as when the model is used in an automated control system in industrial production, where the model must be able to handle various possible abnormal situations and interferences, otherwise it may have catastrophic consequences. In other cases, preventing overfitting may be more important, such as in medical image analysis, where overfitting can lead to incorrect diagnosis and treatment recommendations for patients. In some cases, both robustness and preventing overfitting may be equally important, and a balanced approach is needed to consider both performances. Our method happens to have both of these advantages.

\section{Experiments}

The paper presents experimental results on several classic image datasets, including CIFAR-10 dataset ~\cite{coates2011analysis}, STL-10 dataset~\cite{torralba200880}, CelebA-HQ~\cite{karras2017progressive} . The CIFAR-10 dataset consists of 60K images with a size of (32×32×3). It is divided into 50K training instances and 10K testing instances.

The STL-10 dataset is specifically designed for unsupervised feature learning. It includes 5K training images, 8K testing images, and an additional 100K unlabeled images, all of size (48×48×3).

In addition, our proposed method is also tested on the dataset with higher resolutions: CelebA-HQ . CelebA-HQ contains 30,000 human faces with a size of (256×256×3).

To quantitatively evaluate the effectiveness of our proposed method, we employ two widely recognized evaluation metrics for generative models: Inception Score~\cite{salimans2016improved} and Frechet Inception Distance (FID)~\cite{heusel2017gans}. To ensure a fair comparison, we calculate these metrics using the official implementations, following the standard evaluation protocol.

To provide a comprehensive analysis and compare our results with previous works, which may have deviated from the standard evaluation protocol, we conduct evaluations with different configurations for calculating FID ~\cite{gong2019autogan,kurach2018gan}. This allows us to assess the robustness and performance of our method under various settings. Throughout the training process, we record the model checkpoint that achieves the best FID score and report the averaged results based on this checkpoint.

By adhering to standard evaluation procedures and considering multiple FID configurations, we aim to present a comprehensive and reliable evaluation of our proposed method, enabling meaningful comparisons with previous works in the field.

\subsection{Unconditional Image mage Generation}

The proposed PGN-GAN includes a Wasserstein loss function, but due to the range restriction imposed by the Group Normalization(PGN) layer, the Wasserstein loss degenerates into a hinge loss. Specifically, since the PGN layer constrains the output of the network to be within the range of [-1, 1], the Wasserstein loss function is effectively replaced by a hinge loss function, which penalizes misclassified samples with a linear function. This means that in the context of the PGN-GAN, the original Wasserstein loss function is effectively simplified to a hinge loss function. It is worth noting this relationship between the two loss functions in the PGN-GAN architecture.

Additionally, we employ the Kaiming Normal Initialization technique to initialize the weights of both fully-connected layers and CNN layers in our model~\cite{he2015delving}. This initialization method is known for its effectiveness in deep learning models, as it helps alleviate the vanishing and exploding gradient problems during training. As for the biases, we initialize them to zero. This choice of initialization aims to provide a balanced starting point for the model's parameters, facilitating stable and efficient learning throughout the training process.

\begin{table}[ht]
\centering

\label{tab:results1}
\begin{tabular}{@{}lccc}
\toprule

 Method& \multicolumn{2}{c}{CIFAR-10}  \\

 & Inception Score$\uparrow$ & FID(train)$\downarrow$  \\ 
 \hline
 Real data & 11.24±.12 & 0  \\
 \hline
\textbf{\textit{Standard CNN}} &  &   \\ 
\midrule
SN-GAN~\cite{miyato2018spectral} & 7.58±.12 & -   \\
SN-GAN-CR~\cite{zhang2019consistency}& 7.93 & -   \\
GN-GAN~\cite{wu2021gradient} & 7.71±.14 & 19.31±.76  \\
GN-GAN-CR~\cite{wu2021gradient} & 8.04±.19 & 18.59±1.5 \\
(our)PGN-GAN & \textbf{7.83(0.081)} & 18.675  \\
(our)PGN-GAN-CR & \textbf{8.12(0.062)} & 18.938    \\
\hline
\textbf{\textit{ResNet}} & &   \\
WGAN-GP~\cite{gulrajani2017improved} & 7.86±.08 & -   \\
SN-GAN &  8.22±.05 & -   \\
SN-GAN-CR &8.40  & -   \\
GN-GAN & 8.49±.11 & 11.13±.18 \\
GN-GAN-CR  &8.72±.11 &9.55±.47 \\
(our)PGN-GAN & \textbf{8.506(0.126)} & 11.597   \\
(our)PGN-GAN-CR & \textbf{8.909(0.123)} & \textbf{9.079}   \\
 \hline
 \textbf{\textit{Neural Architecture Search}} & &   \\
 AutoGAN~\cite{gong2019autogan} & 8.55±.10 & 12.42   \\
E²GAN~\cite{tian2020off} & 8.51±.13 & 11.26   \\
 \hline
\end{tabular}
\caption{CNN Benchmark results on Cifar10 dataset.}
\end{table} 

STL-10 and CIFAR-10 are both commonly used datasets consisting of ten different classes of samples. The CIFAR-10 training dataset contains 50,000 images, with an additional validation set of 10,000 images. Since generative tasks benefit from larger amounts of training data, we chose to use the CIFAR-10 training set with 50,000 images as our dataset. In testing our model with the CIFAR-10 dataset, we used 200,000 epochs. We conducted 500,000 runs of the STL-10 training set. For the ResNet architecture~\cite{he2016deep,balduzzi2017shattered}, we slightly increased the learning rate of the discriminator from $2\times10^{-4}$ to $4\times10^{-4}$. Based on the results of the GN-GAN authors' tests, we also chose to set $\lambda=5$ as the experimental parameter for testing CR method with ResNet in PGN-GAN-CR, as it was found to produce better results than other values. We observed from the GN-GAN paper that the choice between hinge loss~\cite{lim2017geometric,miyato2018spectral} and non-saturating loss has little impact on the results. Therefore, for both CIFAR-10 and STL-10 datasets in this study, we used hinge loss and did not conduct separate tests on non-saturating loss or the validation set samples.

The augmentation setting in Table~\ref{cr_pipeline_table_appendix} is used for Consistency Regularization~\cite{zhao2021improved}. 

\renewcommand{\arraystretch}{1.15}
\begin{table}[hbt]
    \centering
    \begin{tabular}{|c|l|}
        \hline
         1. & RandomHorizontalFlipping(p=$0.5$)\\
        \hline
         2. & RandomPixelShifting(pixel=$0.2\times$ImageSize) \\
        \hline
    \end{tabular}
    \caption{Augmentation for consistency regularization on CIFAR-10 and STL-10.}
    \label{cr_pipeline_table_appendix}
\end{table}

It is evident that our PGN method outperforms the GN-GAN-CR method on the STL-10 dataset, even without using the CR method to prevent overfitting. Our PGN method achieves better experimental results on both the CNN and ResNet architectures.

\begin{table}[ht]
\centering

\label{tab:results2}
\begin{tabular}{@{}lccc}
\toprule
 Method& \multicolumn{2}{c}{STL-10}  \\

 & Inception Score$\uparrow$ & FID(train)$\downarrow$  \\ 
 \hline
 Real data & 26.08±.26 & 0  \\
 \hline
Standard CNN  &  &  \\ 
\midrule
SN-GAN~\cite{miyato2018spectral} & 8.79±.14 & 43.20  \\
SN-GAN-CR ~\cite{zhang2019consistency}& 8.69±.08 & 34.14  \\
GN-GAN~\cite{wu2021gradient} & 9.00±.15 & 32.41±.73 \\
GN-GAN-CR~\cite{wu2021gradient} & 9.00±.14 & 27.61±.69 \\
(our)PGN-GAN & \textbf{9.340(0.178)} & \textbf{27.378}    \\
(our)PGN-GAN-CR & \textbf{9.207(0.107)} & \textbf{26.758}   \\
 \hline

ResNet & &   \\
SN-GAN & 9.10±.04 & 40.10±.50  \\
SN-GAN-CR & 9.38±.07 & 28.4  \\
GN-GAN & 9.60±.14 &  26.14±.7 \\
GN-GAN-CR& 9.74±.15 &  23.62±.89 \\
(our)PGN-GAN & \textbf{9.895(0.114)} & \textbf{23.588}  \\
(our)PGN-GAN-CR & \textbf{9.800(0.117)} & \textbf{21.803}    \\
 \hline
 \textbf{\textit{Neural Architecture Search}} & &   \\
 AutoGAN~\cite{gong2019autogan} & 9.16±.12 & 31.01   \\
E²GAN~\cite{tian2020off} & 9.51±.09 & 25.53   \\
 \hline
\end{tabular}
\caption{Benchmark results on STL-10 dataset.}
\end{table}

\subsection{Conditional Image mage Generation}

Furthermore, we adopt the official implementation of BigGAN~\cite{brock2018large} as our baseline. The training process utilizes the Adam optimizer with specific parameter values: $\alpha_G=1\times10^{-4}$, $\alpha_D=2\times10^{-4}$, $\beta_1=0$, $\beta_2=0.999$. We set the batch size to 50 and update the generator once every four discriminator update steps. We terminate the training once the generator reaches 125,000 steps. To augment the real images, we apply random horizontal flipping.

To enhance the stability and performance of the generator, we employ the technique of moving averages on the generator weights, with a decay rate of 0.9999. The conditional refinement (CR) pipeline~\cite{zhao2021improved,mirza2014conditional} we utilize is presented in Table~\ref{cr_pipeline_table_appendix}. In terms of evaluation, Table~\ref{conditional_image_generation_table_appendix} displays the results for different approaches based on the metrics of Inception Score (IS), Fréchet Inception Distance (FID) for the training set. The results suggest that BigGAN with the proposed group penalty normalization (PGN) outperforms BigGAN with CR alone.

\begin{table}[ht]
    \centering
    \caption{Inception Score(IS) and FID of conditional image generation on CIFAR-10.}
    \begin{tabular}{lcc}
        \hline
        Method & IS$\uparrow$ & FID(train)$\downarrow$  \\
        \hline	
        aw-SN-GAN ~\cite{zadorozhnyy2021adaptive} & 9.01±.03 & 8.03 \\
        BigGAN~\cite{brock2018large} & 9.22 & - \\
        GN-BigGAN-CR~\cite{wu2021gradient} & 9.35$\pm$.14 & 4.86$\pm$.07  \\
        (our)PGN-BigGAN-CR & \textbf{9.471(0.119)} & 5.495  \\
        \hline
    \end{tabular}
    \label{conditional_image_generation_table_appendix}
\end{table}

\subsection{Unconditional Image Generation on CelebA-HQ }

We conducted additional experimentation on the CelebA-HQ high-resolution image dataset to further evaluate the effectiveness of the proposed Penalty Gradient Normalization approach. As part of the data augmentation process~\cite{zhao2021improved}, we employed random horizontal flipping for the dataset. In order to generate 256 × 256 images, we utilized the architecture recommended by SN-GAN.

\begin{figure}[hbt!]
    \centering
    \begin{minipage}[t]{0.48\textwidth}
        \centering
        \includegraphics[scale=0.46]{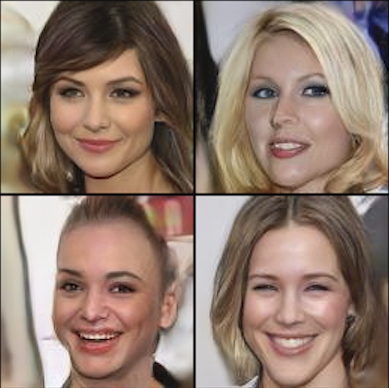}
        \caption{Generated samples on CelebA-HQ 128$\times$128. FID=14.66. (PGN-GAN)}
        \label{large_scale_image_celeba128}
    \end{minipage}\hfill
    \begin{minipage}[t]{0.48\textwidth}
        \centering
        \includegraphics[scale=0.46]{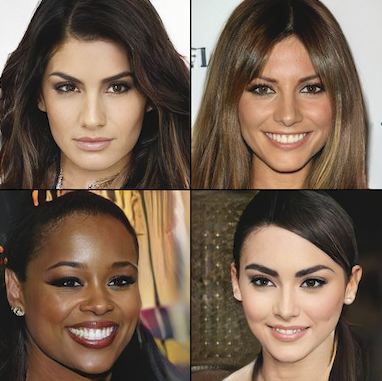}
        \caption{Generated samples on CelebA-HQ 256$\times$256. FID=6.186. (PGN-GAN)}
        \label{large_scale_image_celeba256}
    \end{minipage}
\end{figure}

For optimization, we employed the Adam~\cite{kingma2014adam} optimizer with the following parameter values: $\alpha_G=2\times10^{-4}$, $\alpha_D=2\times10^{-4}$, $\beta_1=0$, $\beta_2=0.9$, and a batch size of 64. The generator was updated once for every 5 discriminator update steps. To ensure stability, we employed a stopping criterion that halted all training processes after the generator had undergone 100k steps. Additionally, we applied moving averages to the generator weights with a decay rate of 0.9999.

The performance of the model was evaluated using two metrics: Inception Score and FID. The results are presented in Table V. It should be noted that by employing a more advanced architecture, it is possible to further enhance the performance of the model.

\begin{table}[ht]
    \centering
    \caption{FID of unconditional image generation on CelebA-HQ. $\dagger$ represents that we provide SN-GAN implementation as the baseline.}
    \begin{tabular}{lccc}
        \hline
        Dataset     & PGN-GAN    & GN-GAN     & SN-GAN            \\
        \hline
        CelebA-HQ 128  & 14.66  & 14.78 & 25.95(from~\cite{patel2021lt})   \\
        CelebA-HQ 256  &\textbf{6.186} & 7.67 & 14.45$^\dagger$   \\
        \hline
    \end{tabular}
    \label{large_scale_unconditional_image_generation_table_appendix}

\end{table}

\section{ Conclusion and Future Work}

In this paper, we present a new method for Penalty normalizing gradients that can be used to improve the training of GANs, and which has the potential to enhance a variety of applications. Our method, known as PGN, is straightforward to implement and has been mathematically proven to generate gradients with bounded norms. We have applied PGN to different GAN architectures on a range of datasets, and achieved state-of-the-art results for most of them.

One of the main advantages of Penalty Gradient Normalization is that it is computationally efficient. Unlike other normalization techniques, such as batch normalization, it does not require the computation of running statistics or the calculation of additional parameters. This makes it well-suited for use in large, complex GAN models where computational efficiency is a concern.

Another advantage of Penalty Gradient Normalization is that it can be easily implemented in existing GAN architectures without requiring major modifications. This makes it a flexible and practical solution for improving the performance of GAN models.

In summary, Penalty Gradient Normalization is a powerful normalization technique for GANs that helps to stabilize the training process, improve the quality of the generated samples, and is computationally efficient.

Regarding future work, we propose the idea of automating the process of finding better formulas for optimizing GAN performance. This could involve using artificial intelligence to test and verify alternative formulas for various mathematical operations, including cubes, quartiles, multiplication, division, or even the sin function. Once our current research is complete, we believe that it would be valuable to pursue such a more advanced approach.

\section*{Acknowledgments}
 The authors acknowledge support in the form of ``Support for Pioneering Research Initiated by the Next Generation (SPRING)'' from the Japan Science and Technology Agency. This work was supported by JST SPRING, Grant Number JPMJSP2114. 
  \\
\newline
Computations were partially performed on the NIG supercomputer at ROIS National Institute of Genetics.
 \\
\newline
 We would like to express our sincere gratitude to Mr. Yilun Wu, a PhD candidate at National Yang Ming Chiao Tung University, Taiwan, for his valuable support and guidance throughout our research work.

\clearpage
\section*{A. Theoretical Results}

The mathematical formula derivation in this paper is also largely based on GN-GAN~\cite{wu2021gradient}. 
We first define the Lipschitz constant $L_f$ again for better readability. Let $f:\mathbb{R}^n \rightarrow \mathbb{R}$ be a mapping function. Then, $L_f$ is the minimum real number such that:

\begin{equation}
    \begin{aligned}
        \vert f(x)-f(y) \vert\le L_f\Vert x - y\Vert,\forall x,y\in\mathbb{R}^n.
    \end{aligned}
    \label{lipschitz_constraints_appendix}
\end{equation}

Lemma 3. Let $f:\mathbb{R}^n \rightarrow \mathbb{R}$ be a continuously differentiable function and $L_f$ be the Lipschitz constant of $f$. Then, the Lipschitz constraint (14) is equivalent to
\begin{equation}
    \begin{aligned}
        \Vert\nabla_x f(x)\Vert\le L_f,\forall x\in\mathbb{R}^n.
    \end{aligned}
    \label{grad_constraint_appendix}
\end{equation}

Proof. We first prove the sufficient condition. $(\Rightarrow)$ From the definition of Lipschitz constraint (14), we know

\begin{equation}
    \vert f(x)-f(y)\vert\le L_f\Vert x-y\Vert.
    \end{equation}

Now, we consider the norm of directional derivative at $x$ along with the direction of $(y-x)$:
    \begin{equation}
        \langle\nabla f(x),\frac{y-x}{\Vert y-x\Vert}\rangle=\lim_{y\rightarrow x}\frac{\vert f(y)-f(x)\vert}{\Vert x-y\Vert}\le L_{f},
    \end{equation}
Since the norm of the gradient is the maximum norm of directional derivative, then

    \begin{equation}
        \Vert\nabla f(x)\Vert\le L_{f}.
        \label{grad_constraint}
    \end{equation}
    
We then prove the necessary condition. $(\Leftarrow)$ By the assumption, $f$ is continuous and differentiable. Therefore, the conditions of Gradient theorem are satisfied, and thus we can only consider the line integral along the straight line from $y$ to $x$:

\begin{subequations}
\begin{align}
\vert f(x)-f(y)\vert \\
&=\Big\vert\int_y^x\nabla f(r) dr\Big\vert \\
&=\Big\vert\int_0^1\langle \nabla f(xt+y(1-t)),x-y\rangle dt\Big\vert \\
&\le\Big\vert\int_0^1\Vert \nabla f(xt+y(1-t))\Vert\cdot\Vert x-y\Vert dt\Big\vert \\
&\le L_{f} \Big\vert\int_0^1\Vert x-y\Vert dt\Big\vert \\
&=L_{f}\Vert x-y\Vert.
\end{align}
\end{subequations}

The theorem follows.

Theorem 5. Let $f_K:\mathbb{R}^n\rightarrow\mathbb{R}$ be a layer-wise 1-Lipschitz constrained $K$-layer network. The Lipschitz constant of the first $k$-layer network $L_{f_k}$ is upper-bounded by $L_{f_{k-1}}$, i.e.,

\begin{equation}
    L_{f_k}\le L_{f_{k-1}},\forall k\in \{2\cdots K\}.
\end{equation}

Proof. Since all the layers, including activation functions, are 1-Lipschitz constrained, i.e.,
\begin{equation}
\|W_kx-W_ky\| \leq \|x-y\|,\quad \forall x,y\in\mathbb{R}\
\end{equation}

we have by the triangle inequality,

\begin{equation}
\begin{aligned}
&\Vert f_k(x)-f_k(y) \Vert \\
& = |\varphi_k(W_kf_{k-1}(x)+b_k)- \\
& \quad \varphi_k(W_kf_{k-1}(y)+b_k)| \\
& \leq L_{\varphi_k}|W_kf_{k-1}(x)- \\
& \quad W_kf_{k-1}(y)| \\
& \leq L_{\varphi_k}L_{f_{k-1}}|f_{k-1}(x)- \\
& \quad f_{k-1}(y)|.
\end{aligned}
\end{equation}

where the last equality follows from the 1-Lipschitz constraint on the activation function $\varphi_{k-1}$. 
This completes the proof.

The penalty gradient norm bound for the modified normalized function $\hat{f}(x)=(1-f(x))/\big(\Vert 1 - \nabla_x f(x)\Vert+\vert 1 - f(x)\vert\big)$ can be obtained in a similar way as in the original gradient norm proof in GN-GAN. 

We first use the fact that $f$ is bounded between 0 and 1, we apply the following inequality:

\begin{equation}
\label{normalized_gradient_norm_appendix_proof3}
\frac{\Vert\nabla f\Vert^2}{(\Vert\nabla f\Vert + \vert f\vert)^2} \leq 1, \quad\text{for}\quad \Vert\nabla f\Vert, \vert f\vert \geq 0.
\end{equation}

We can substitute $f$ with $1-f$ since we have used the fact that $f$ is bounded between 0 and 1 in this proof.\\
\noindent\textbf{Theorem 6.}
\textit{Let $f:\mathbb{R}^n\rightarrow\mathbb{R}$ be a continuous function which is modeled by a neural network, and all the activation functions of network $f$ are piecewise linear. The modified normalized function $\hat{f}(x)=(1-f(x))/\big(\lVert 1 - \nabla_x f(x)\rVert+\lvert 1 - f(x)\rvert\big)$ is penalty gradient norm bounded, i.e.,}

\begin{equation}
\Vert\nabla_x\hat{f}(x)\Vert=\Bigg\Vert\frac{\Vert\nabla (1-f)\Vert}{\Vert\nabla (1-f)\Vert+\vert 1 - f\vert}\Bigg\Vert^2\le 1.
\end{equation}

First, we can derive the denominator part.
\begin{equation}
\begin{aligned}
\left(\Vert\nabla f\Vert + \vert f \vert\right)^2 &= \Vert\nabla f\Vert^2 + 2\Vert\nabla f\Vert \vert f \vert + \vert f \vert^2 \\
&= \Vert\nabla(-1+f)\Vert^2 + 2\Vert\nabla(-1+f)\Vert \vert -1+f \vert \\
&\quad + (-1+f)^2 \\
&\ge \Vert\nabla(-1+f)\Vert^2 + 2\Vert\nabla(-1+f)\Vert \vert -1+f \vert \\
&= \Vert\nabla(1-f)\Vert \left(\Vert\nabla(1-f)\Vert + 2\vert 1-f \vert\right) \\
&= \Vert\nabla(1-f)\Vert^2 + 2\Vert\nabla(1-f)\Vert \vert 1-f \vert + \vert 1-f \vert^2 \\
&= \Vert\nabla(-f)\Vert^2 + 2\Vert\nabla(-f)\Vert \vert -f \vert + (-f)^2 \\
&= \left(\Vert\nabla(-f)\Vert + \vert -f \vert\right)^2 \\
&= \left(\Vert\nabla(1-f)\Vert + \vert 1-f \vert\right)^2 
\end{aligned}
\end{equation}

Next, let's work on the numerator.

\begin{equation}
\begin{aligned}
\Vert\nabla(1-f)\Vert^2 &= \langle\nabla(1-f), \nabla(1-f)\rangle \\
&= \sum_i \left(\frac{\partial(1-f)}{\partial x_i}\right)^2 \\
&= \sum_i (-\frac{\partial f}{\partial x_i})^2 \\
&= \Vert\nabla f\Vert^2.
\end{aligned}
\end{equation}

\begin{proof}

By definition, the gradient norm of $\hat{f}(x)$ is:

\begin{subequations}
\begin{align}
\Vert\nabla\hat{f}\Vert
&=\Bigg\Vert\nabla\bigg(\frac{1-f}{\Vert 1 - \nabla f\Vert+\vert 1 - f\vert}\bigg)\Bigg\Vert \\
&=\Bigg\Vert\frac{\nabla (1-f)\big(\Vert 1 - \nabla f\Vert+\vert 1 - f\vert\big)}{\big(\Vert 1 -
\nabla f\Vert+\vert 1 - f\vert\big)^2} \nonumber \\
&\qquad - \frac{(1-f)\nabla\big(\Vert 1 - \nabla f\Vert+\vert 1 - f\vert\big)}{\big(\Vert 1 -
\nabla f\Vert+\vert 1 - f\vert\big)^2}\Bigg\Vert.
\label{th6:pfeq2}
\end{align}
\end{subequations}

By simple chain rule, we know that:
\begin{subequations}
\begin{align}
\nabla\Vert\nabla (1-f)\Vert&=\nabla^2 (1-f)\frac{\nabla (1-f)}{\Vert\nabla (1-f)\Vert}, \\
\nabla\vert 1-f\vert&=\nabla (1-f)\frac{1-f}{\vert 1-f\vert}.
\end{align}
\end{subequations}

Since the network $f$ contains only piecewise linear activation functions, the Hessian matrix $\nabla^2 f$ is a zero matrix. The Eq.\eqref{th6:pfeq2} can be simplified:

\begin{equation}
\label{modified_normalized_gradient_norm_appendix}
\begin{aligned}
\Vert\nabla\hat{f}\Vert
&=\Bigg\Vert\frac{\Vert\nabla (1-f)\Vert^2}{\big(\Vert\nabla (1-f)\Vert+\vert 1 - f\vert\big)^2}\Bigg\Vert \\
&=\Bigg\Vert\frac{\Vert\nabla (1-f)\Vert}{\Vert\nabla (1-f)\Vert+\vert 1 - f\vert}\Bigg\Vert^2\le 1.\\
\end{aligned}
\end{equation}
The theorem follows.

\end{proof}

\clearpage

\begin{figure*}[hbth!]%
    \centering
    
    \subfigure[PGN-GAN CNN]{%
        \includegraphics[width=0.48\textwidth]{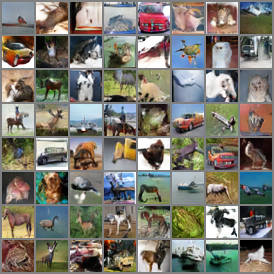}
        \label{subfig:pgn_gan_cnn_cifar10}}
    \hfill
    \subfigure[PGN-GAN-CR CNN]{%
        \includegraphics[width=0.48\textwidth]{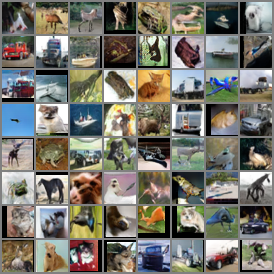}
        \label{subfig:pgn_gan_cr_cnn_cifar10}}
    
    \vspace{0.5cm}
    
    \subfigure[PGN-GAN ResNet]{%
        \includegraphics[width=0.48\textwidth]{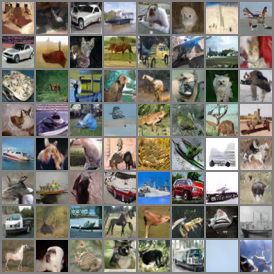}
        \label{subfig:pgn_gan_resnet_cifar10}}
    \hfill
    \subfigure[PGN-GAN-CR ResNet]{%
        \includegraphics[width=0.48\textwidth]{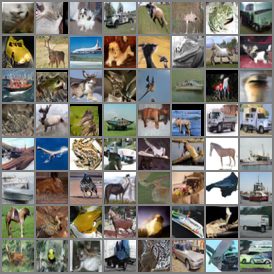}
        \label{subfig:pgn_gan_cr_resnet_cifar10}}
    
    \caption{Unconditional image generation on CIFAR-10.}
    \label{fig:cifar_10_images}
\end{figure*}

\begin{figure*}[h!]
    \centering
    
    \subfigure[PGN-GAN CNN]{%
        \includegraphics[width=0.48\textwidth]{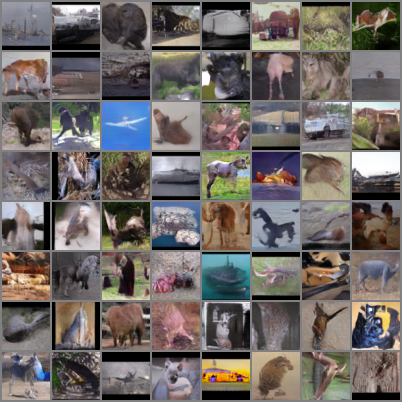}
        \label{subfig:pgn_gan_cnn_stl10}}
    \hfill
    \subfigure[PGN-GAN-CR CNN]{%
        \includegraphics[width=0.48\textwidth]{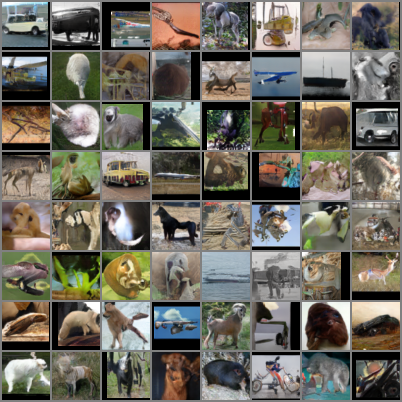}
        \label{subfig:pgn_gan_cr_cnn_stl10}}
    
    \vspace{0.5cm}
    
    \subfigure[PGN-GAN ResNet]{%
        \includegraphics[width=0.48\textwidth]{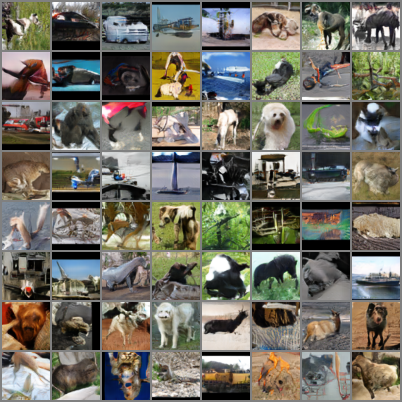}
        \label{subfig:pgn_gan_resnet_stl10}}
    \hfill
    \subfigure[PGN-GAN-CR ResNet]{%
        \includegraphics[width=0.48\textwidth]{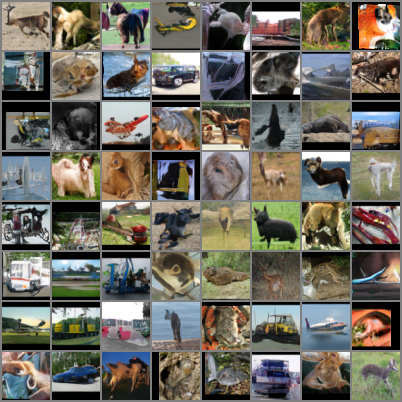}
        \label{subfig:pgn_gan_cr_resnet_stl10}}
    
    \caption{Unconditional image generation on STL-10.}
    \label{unconditional_stl_10_figure_appendix}
\end{figure*}

\begin{figure*}[p]
    \centering
    \includegraphics[scale=0.40]{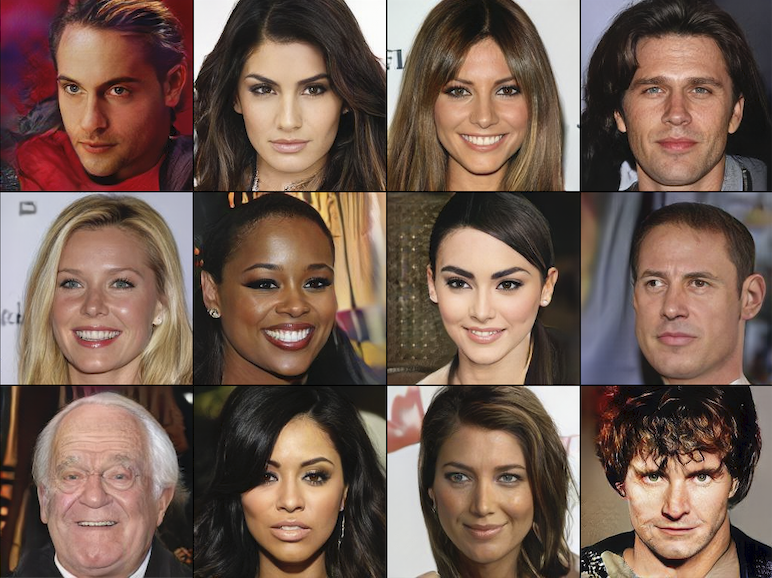}
    \caption{PGN Unconditional image generation on CelebA-HQ $256\times256$.}
    \label{celebhq_7x7_figure_appendix_a}
\end{figure*} 

\begin{figure*}[p]
    \centering
    \includegraphics[scale=0.40]{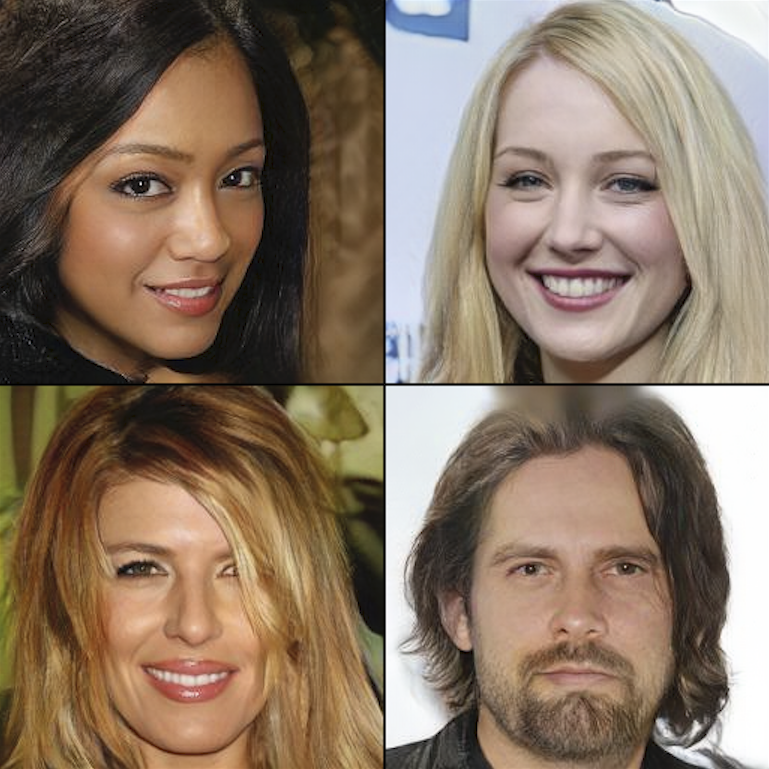}
    \caption{PGN Unconditional image generation on CelebA-HQ $256\times256$.}
    \label{celebhq_7x7_figure_appendix_b}
\end{figure*}

\begin{figure*}[p]
    \centering
    
    \subfigure[sample]{%
        \includegraphics[scale=0.50]{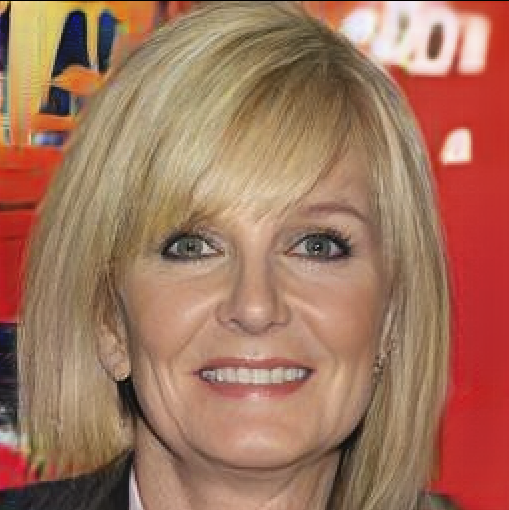}}
    \hfill
    \subfigure[sample]{%
        \includegraphics[scale=0.50]{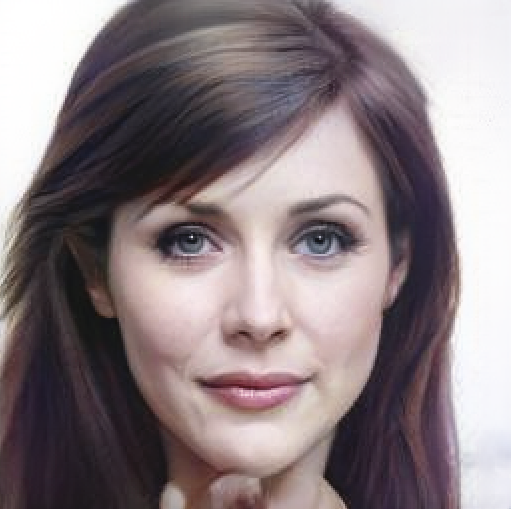}}   
    
    \caption{PGN Unconditional image generation on CelebA-HQ $256\times256$.}
    \label{celebhq_7x7_figure_appendix_c}
\end{figure*}

\begin{figure*}[p]
    \centering
        
    \subfigure[sample]{%
        \includegraphics[scale=0.50]{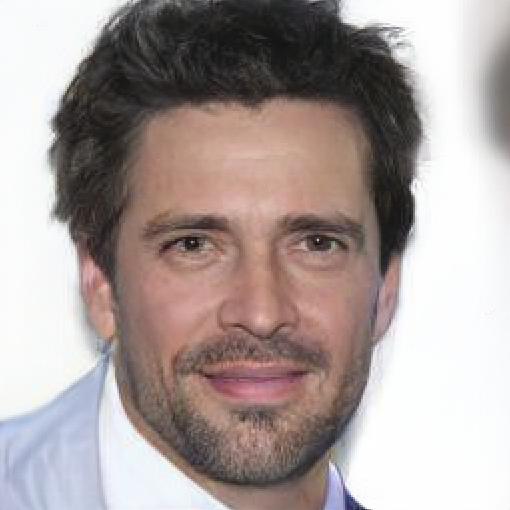}}
    \hfill
    \subfigure[sample]{%
        \includegraphics[scale=0.50]{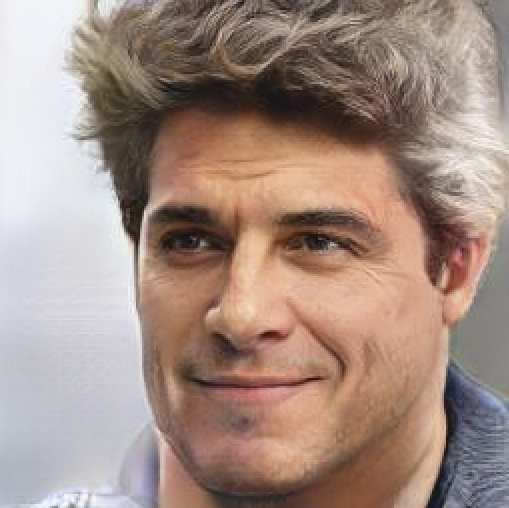}}
        
    \caption{PGN Unconditional image generation on CelebA-HQ $256\times256$.}
    \label{celebhq_7x7_figure_appendix_d}
\end{figure*}

\clearpage

\bibliographystyle{IEEEtran}
\bibliography{bibliography.bib}

\addtolength{\textheight}{-12cm}   



\end{document}